\documentclass{article}

\usepackage{arxiv}

\usepackage[utf8]{inputenc} 
\usepackage[T1]{fontenc}    
\usepackage{url}            
\usepackage{booktabs}       
\usepackage{amsfonts}       
\usepackage{nicefrac}       
\usepackage{microtype}      
\usepackage{lipsum}
\usepackage{float}

\usepackage{cite}
\usepackage{amsmath,amssymb,amsfonts}
\usepackage{algorithm}
\usepackage{algpseudocode}
\usepackage{graphicx}
\usepackage{textcomp}
\usepackage{bmpsize}
\usepackage[dvipsnames,table,xcdraw]{xcolor}

\usepackage{lipsum}
\usepackage[colorlinks=true,urlcolor=black]{hyperref}
\usepackage{mathtools}
\usepackage[numbers]{natbib}
\usepackage{subcaption}

\usepackage{multirow}
\usepackage{url}
\usepackage{booktabs}
\usepackage{amsfonts} 
\usepackage{nicefrac} 
\usepackage{microtype}
\usepackage{adjustbox}
\usepackage{todonotes}
\usepackage{amsmath}
\usepackage{cleveref}
\usepackage{xcolor}

\usepackage{amsfonts}       

\usepackage{lipsum}
\usepackage{comment}
\usepackage{tabularx}
\newcolumntype{Y}{>{\centering\arraybackslash}X}

\definecolor{mysidenotefgcolor}{rgb}{0,0,0}
\definecolor{myorange}{rgb}{1,0.7,0.3}
\definecolor{mysidenotefgauthorcolor}{rgb}{0.5,0.5,0.5}

\title{Efficient Adversarial Training with Data Pruning}

\let\oldnl\nl
\newcommand{\nonl}{\renewcommand{\nl}{\let\nl\oldnl}}

\author{
  Maximillian Kaufmann \\
  University of Cambridge \\
  \And
  Yiren Zhao \\
  University of Cambridge\\
  \And
  Ilia Shumailov \\
  University of Cambridge and Vector Institute\\
  \And 
  Robert Mullins \\
  University of Cambridge\\
  \And
  Nicolas Papernot \\
  University of Toronto and Vector Institute \\
}

\begin{document}
\maketitle

\begin{abstract}
Neural networks are susceptible to  adversarial examples --- small input perturbations that cause models to fail. Adversarial training is one of the solutions that stops adversarial examples; models are exposed to attacks during training and learn to be resilient to them. Yet, such a procedure is currently expensive -- it takes a long time to produce and train models with adversarial samples, and, what is worse, it occasionally fails. In this paper we demonstrate \textit{data pruning} --- a method for increasing adversarial training efficiency through data sub-sampling. We empirically show that \textit{data pruning} leads to improvements in convergence and reliability of adversarial training, albeit with different levels of utility degradation. For example, we observe that using random sub-sampling of CIFAR10 to drop 40\% of data, we lose 8\% adversarial accuracy against the strongest attackers, while by using only 20\% of data we lose 14\% adversarial accuracy and reduce runtime by  a factor of $3$. Interestingly, we discover that in some settings data pruning brings benefits from both worlds --- it both improves adversarial accuracy and training time. 
\end{abstract}
\section{Introduction}

The existence of \emph{adversarial examples}~\cite{goodfellow2014explaining} poses a real threat to Machine Learning (ML) systems. These are an interesting phenomenon in modern neural networks, wherein small (semantic preserving) perturbations to network inputs lead to incorrect behaviour. Although first found in the image
classification context \cite{szegedy2013intriguing,biggio2013evasion}, they have since been discovered in a variety of domains---including
natural language processing~\cite{samanta2017crafting,boucher2021bad}, audio processing~\cite{carlini2018audio,ahmed2022pipe} 
and deep reinforcement learning~\cite{gleave2021adversarial,zhao2020blackbox}. 

A common technique to make models less susceptible to such attacks is \emph{adversarial training}~\cite{goodfellow2015explaining}, wherein correctly labelled adversarial examples are shown at training time, allowing the models to learn  correct behaviour even in the face of malicious inputs -- such models are said to be \emph{adversarially robust}. In the image classification context, training of this form has been successful in decreasing the susceptibility of models to adversarial examples~\cite{goodfellow2014explaining,madry2019deep,fast_github,evaluatingrobustness}. However, traditional adversarial training regimes have often been prohibitively expensive, with adversarial training being up to thirty times more computationally expensive than training a non-robust model~\cite{shafahi2019free}. Historically, this meant that training robust models on large datasets has been exclusive to research groups with access to hundreds of GPUs~\cite{xie2019feature,kannan2018adversarial}. Furthermore, the high computational cost adds extra barriers to these systems being deployed in practice---this is relevant when viewing adversarial robustness as an important safety property for modern systems \cite{hendrycks2021unsolved},

In this paper we seek to explore and expand on these methods---with the aim of continuing to reduce the barrier to training robust models. To this end, we propose and design a class of \emph{Data Pruning} techniques with different complexities for adversarial training, with the goal of reducing its overall runtime. By sub-sampling the training dataset, we observe  significant reduction in adversarial training time, while balancing standard and adversarial performance. We find that depending on what data is omitted, it is possible to favour adversarial performance and design more complex regimes for adversarial training. We also demonstrate relatively unexpected results---in cases where lack of data is not a constraint to model performance, we observe that dropping more data leads to increased adversarial robustness. Our work suggest that in the future a more advanced sub-sampling strategy may help improve adversarial robustness, while providing significant speed ups from selecting information-efficient data subsets.

Overall we make the following contributions:

\begin{itemize}
    \item We design a class of datapoint loss-value based \textit{Data Pruning} techniques for reducing the run-time of adversarial training. These methods employ different dataset sub-sampling strategies to reduce the number of data points required for adversarial training, while aiming to keep robustness high.
    \item We show that simple uniform random sub-sampling can significantly reduce the run-time with a certain robustness accuracy drop---dropping 40\% of data leads to a 30\% runtime reduction, while reducing robustness by 7\% . In addition, we observe that the pruning of high-loss data-points is an effective strategy on MNIST---finding we can drop 60\% of data while increasing the robustness of models.
    \item We evaluate data pruning with different sub-sampling strategies, and report their AutoAttack accuracy with recorded training time on a set of standard benchmarks. 
\end{itemize}

\section{Related work}

\subsection{Adversarial Machine Learning}
Adversarial machine learning in the context of modern deep networks was introduced in a seminal papers by \emph{Szegedy et al.}~\cite{szegedy2013intriguing} and \emph{Biggio et al.}~\cite{biggio2013evasion}, where it was noted that modern machine learning approaches are vulnerable when faced with small perturbations to their inputs. A follow up paper by \emph{Goodfellow et al.} \cite{goodfellow2014explaining} then gave the first example of \emph{adversarial training}, introducing a relatively simple
\emph{Fast Gradient Sign Method} (FGSM) for the generation of adversarial examples. This was the paper originally noting that exposing models to adversarial examples at training time increases their performance in the face of an attacker. 

A large academic community has formed around the study of adversarial examples \cite {shield,chakraborty2018adversarial,bugsfeatures,zhao2020blackbox}, with much `back and forth' between the creation \cite{tao2018attacks,shield,def_dill,detect} and subsequent breaking \cite{ami_broke,shield_broke,detect_broke,def_dill_broke,gao2022limitations,shumailov2018taboo,shumailov2019sitatapatra} of potential defence strategies. Throughout this evolution, adversarial training has remained an effective method for creating robust neural networks. It was originally believed that robust adversarial training was only possible with `strong' attack such as \emph{Projected Gradient Descent} (PGD)~\cite{madry2019deep}, which are generally computationally expensive. More recent work~\cite{wong2020fast} shows that `weaker' attacks, such as the \emph{Random Start-Fast Gradient Sign Method} (RS-FGSM), can still create robust models. 

There is much literature on evaluating the robustness of neural networks~\cite{evaluatingrobustness,evaluating_1,formal_continous_safety_verification,evaluating_3}. Throughout this work, we use the state-of-the-art AutoAttack library~\cite{autoattack}, which provides high-quality, parameter-free attacks for the validation of the robustness of our trained networks.

\section{Methodology}

\subsection{The computational cost of adversarial training}
\label{sec:computational_cost}
The computational cost of adversarial training is very dependant on the exact algorithms used, where performance can be factored as follows. In a single one of the $E$ epochs $\mathcal{O}(N)$ gradient operations occur on each of the $M$ data points. Since training performance is dominated by the number of passes required through the network~\cite{wong2020fast}, this factors the run-time to be well described by $\mathcal{O}(NME)$. This points towards three directions for improvements in computational-efficiency:
\begin{enumerate}
    \item \textbf{Reduce $N$ (Number of network passes)}: As is done in the RS+FGSM algorithm, reducing the number of passes required to compute an adversarial example (while preserving its effectiveness) decreases computation time. Since our method is developed on top of RS+FGSM, we exploit this reduction.
    \item \textbf{Reduce $M$ (Number of data points)}: One can aim to reduce the amount of data required for convergence of the algorithm, by carefully selecting datapoints.
    \item \textbf{Reduce $E$ (Number of epochs)}: Faster convergence of the algorithm leads to reduced computation time.
\end{enumerate}

When aiming to further reduce the computational costs of adversarial training, these factors can be separately considered. Due to the existing low-cost nature of the RS+FGSM (only one gradient pass), and the fact that faster convergence of neural networks is already heavily optimised in the modern machine learning pipeline, \textbf{we focus instead on increasing performance by training on less data}.

\subsection{Data pruning}
\label{sec:method:data_prune}

To reduce the number of data points, we consider subsampling the data, a process controlled by two hyperparemeters:

\begin{itemize}
    \item \textbf{Pruning Epoch} ($e$): This decides at which training epoch to dropout the data
    \item \textbf{Pruning proportion} ($p$): This decides what proportion of data is removed at the pruning epoch.
 \end{itemize}
 
Within this work, we explore the following forms of subsampling:

\textbf{Random data sub-sampling}: This is a simple algorithm, where on a particular epoch a random subset of data is removed. The intuition here is that most datasets contain redundant~\cite{redundancy_data} and incorrectly labeled data~\cite{northcutt2021pervasive}, and hence indiscriminately dropping data may not harm performance significantly and can actually help.

\textbf{Loss-based data sub-sampling}: A more complex data pruning strategy, here the sub-sampling is done based on the loss values of data points given by $\mathcal{L}(f_\theta(\textbf{x}),y)$, where $\mathcal{L}$ is the loss function, $\theta$ the model parameters and $x$ and $y$ are inputs and outputs respectively.

The aim is to have a smaller dataset, while preserving the semantic depth of the original dataset as much as is possible---in the literature this problem is called finding a `core-set', which can be trained on to achieve similar results as on the full dataset~\cite{core-set}. These two training strategies are motivated by separate intuitions:
\begin{itemize}
\item \textbf{High-loss data points}: As described in detail within \Cref{discussion}, high-loss data points are those in which non-robust features are more highly predictive, so can be dropped out with minimal effect on adversarial accuracy. Furthermore, mislabelled examples are likely to have high loss, and as visualised in \Cref{fig:pruning_visualistion}, disproportionately affect robustness.

\item \textbf{Low-loss data points}: These are data points which the model is highly confident on, and likely express patterns which it has seen many times throughout training. Intuitively, it seems likely that some of these points express redundant information, and are within the dense regions of \Cref{fig:clean_example}. However, as shown in \Cref{Evaluation} and explained in \Cref{discussion}, this intuition does not hold in practice.
\end{itemize}


\begin{algorithm}[t]
\caption{RS+FGSM based adversarial training of a classifier $f_\theta$ for $E$ epochs, on a dataset $S$ of size $M$, using learning rate $\eta$, subsampling method $F$, drop proportion $d$. The adversary here is constrained to the $l_\infty$ $\epsilon$-ball, and uses a step size of $\alpha$.} \label{rsfgsm_training_algorithm}
\begin{algorithmic}[1]
\Procedure{RsFfgsmTraining}{$f_\theta$,$S$}
\For{$e = 1\ldots E$}


\If{e == pe}
    \State M' = \{\}
    \For{$i=1\ldots M$}
        \State $M'$.\Call{add}{($i$, $\ell(f_\theta(x_i + \delta),y_i)$)}
        
        \State \Call{sort}{M'} \Comment{using loss magnitude to sort data}
        
        \If{F == low}
            \State Drop $d*|M|$ data from M' with low loss
        \EndIf
        
        \If{F == high}
            \State Drop $d*|M|$ data from M' with high loss
        \EndIf
        
        \If{F == random}
            \State Drop $d*|M|$ data from M' with probability $\frac{1}{|M|}$
        \EndIf
        
        \If{F == high+low}
            \State \Call{sort}{M'} \Comment{using abs(loss magnitude - mean loss magnitude) to sort data}
            \State Drop $d*|M|$ data with lowest cost
        \EndIf
        
    \EndFor
    \State{$S \gets S$}
\EndIf

\For{$i=1\ldots M$}
\State $x_i,y_i \gets \Call{GetDatapoint}{i, S}$ 
\State $\delta \gets \Call{UniformRv}{-\epsilon,\epsilon}$ \Comment{element-wise uniform initialisation in the $l_\infty$ ball}\label{gen_delta_fgsm}
\State$\delta \gets \delta + \alpha\operatorname{sgn}(\nabla_\delta\mathcal{L}(f_\theta(x_i + \delta),y_i))$ \Comment{gradient descent to find adversarial example}
\State $\delta \gets \operatorname{max}(\operatorname{min}(\delta,\epsilon), -\epsilon)$ \Comment{element-wise clipping to keep $\delta$ in $l_\infty$ ball}
\State $\theta \gets \theta - \eta\nabla_\theta \ell(f_\theta(x_i),y_i)$ 
\EndFor
\EndFor
\EndProcedure
\end{algorithmic}
\end{algorithm}
\section{Evaluation}
\label{Evaluation}
\subsection{Experiment setup}
In our experiments, we consider two datasets, MNIST~\cite{lecun2010mnist} and CIFAR10~\cite{krizhevsky2009learning}, and use the AutoAttack~\cite{autoattack} library to produce adversarial examples. To be compatible with~\cite{wong2020fast}\footnote{Definitions can be found in~\url{https://github.com/locuslab/fast_adversarial}.}, we use a standard LeNet-like network for MNIST and PreAct-ResNet18~\cite{he2015deep} for CIFAR10, these networks achieve 99\% and 91\% standard accuracy respectively when trained without the use of adversarial methods. Since our objective is to evaluate the effect of data pruning, we benchmark it without using mixed-precision and early stopping~\cite{wong2020fast}. This means our models use FGSM adversarial training that has the same parameter setup as \textit{Wong et al.}.

\subsection{The effect of the pruning epoch}
In our setup, we pick a specific epoch and trim the dataset and this data dropout \textit{happens only once during the whole adversarial training process}.
There is a tension between runtime efficiency and performance---earlier epochs are better for efficiency, since the model is trained with a full dataset for a shorter time, but dropping out too early could cause issues in model performance. 

There is evidence in the literature that networks do most of their convergence in the first epoch, and that they proceed to forget features of data which they stop seeing at some point during training \cite{forgetting}. Together, these results imply that any data that is dropped out is forgotten (so it does not matter when it is dropped out), and that the decision of which data to drop out will not change after the early epochs (as the model has already mostly converged). Therefore, as early data removal is preferred for efficiency reasons, this would imply that data should be pruned as early as possible. 

\begin{figure}[!h]
	\centering
	\begin{subfigure}{.45\textwidth}
		\centering
		\includegraphics[scale=0.4]{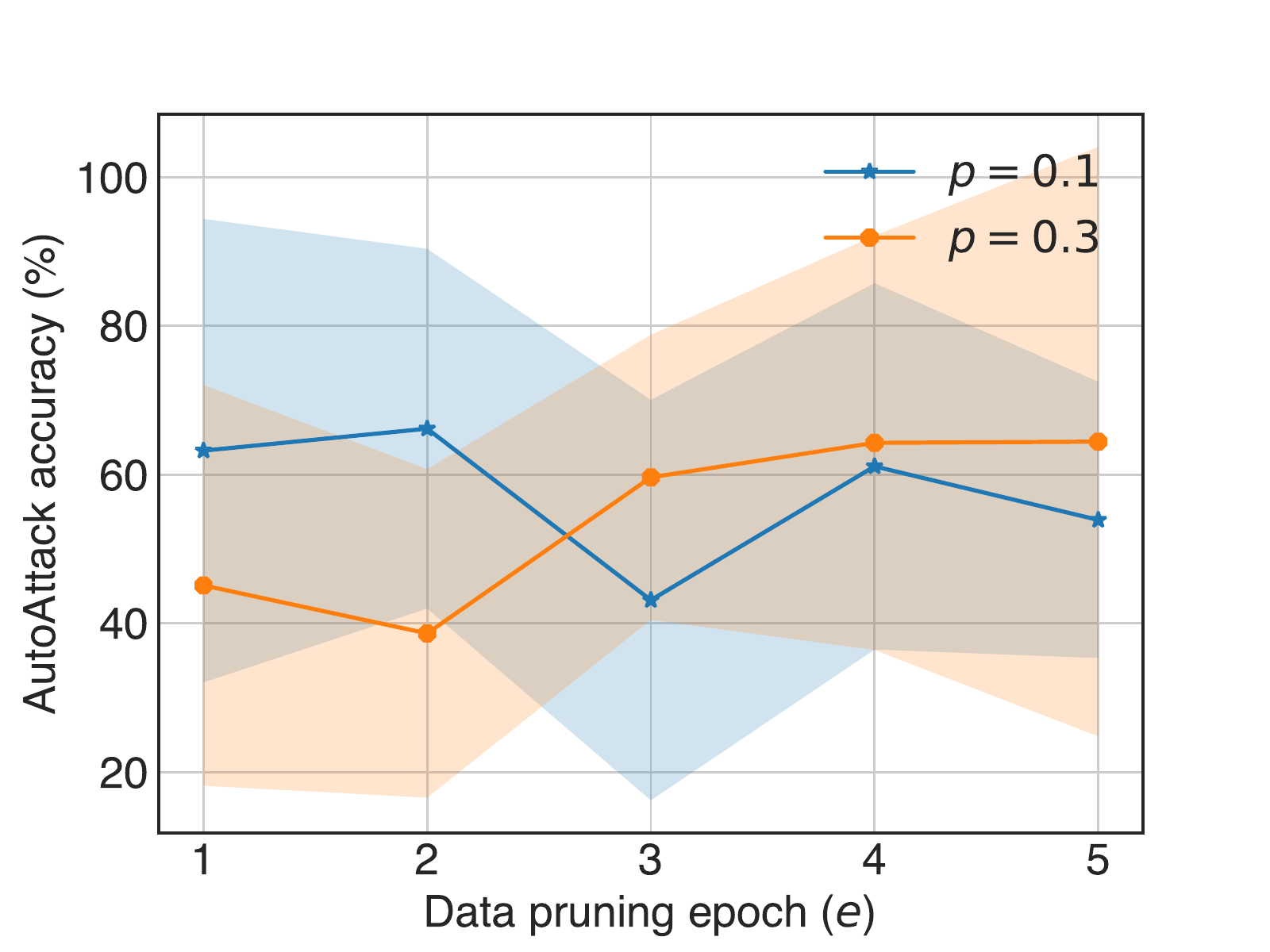}
		\caption{MNIST}
	\end{subfigure}%
	\qquad
	\begin{subfigure}{.45\textwidth}
		\centering
		\includegraphics[scale=0.4]{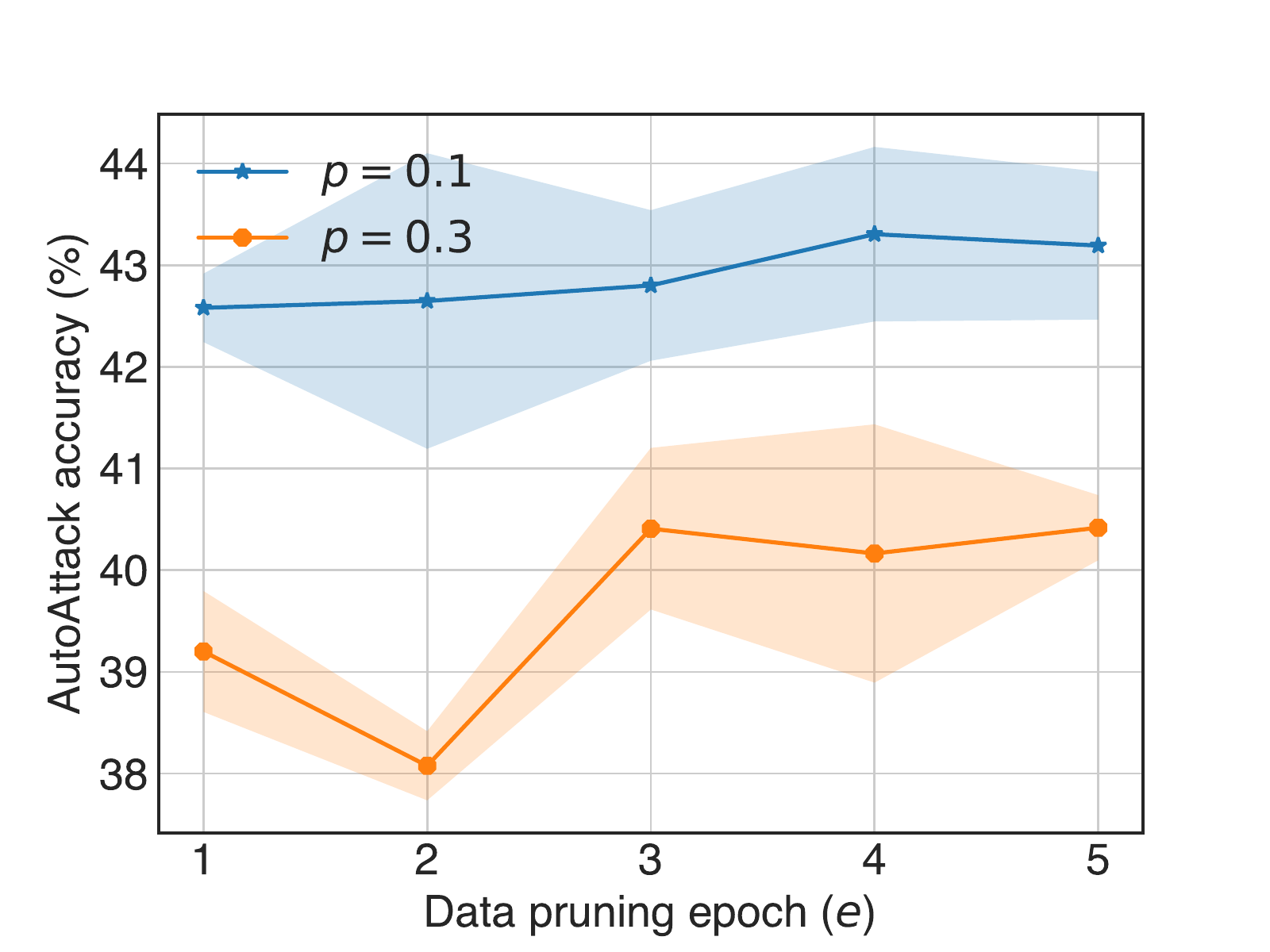}
		\caption{CIFAR10}
	\end{subfigure}
	\caption{Adversarial accuracy against data pruning epoch, for two fixed values of $p$ (0.1 and 0.3 respectively) using \textit{random pruning}, on each of MNIST and CIFAR10. Error bars indicate one standard deviation, and 5 models are trained per data point.}
	\label{fig:eval_random_dropout_search}
	\end{figure}

We investigate this claim empirically, by showing the effect of randomly pruning data at different epochs of adversarial training in \Cref{fig:eval_random_dropout_search}. Given this data, we settle on a dropout epoch of $3$.

\textbf{\textit{Takeaway}}: \textit{Data pruning at a late epoch does not significantly improve performance of the model.}

\subsection{Pruning data with different proportions}
\begin{figure}[!h]
	\centering
	\begin{subfigure}{.45\textwidth}
		\centering
		\includegraphics[scale=0.4]{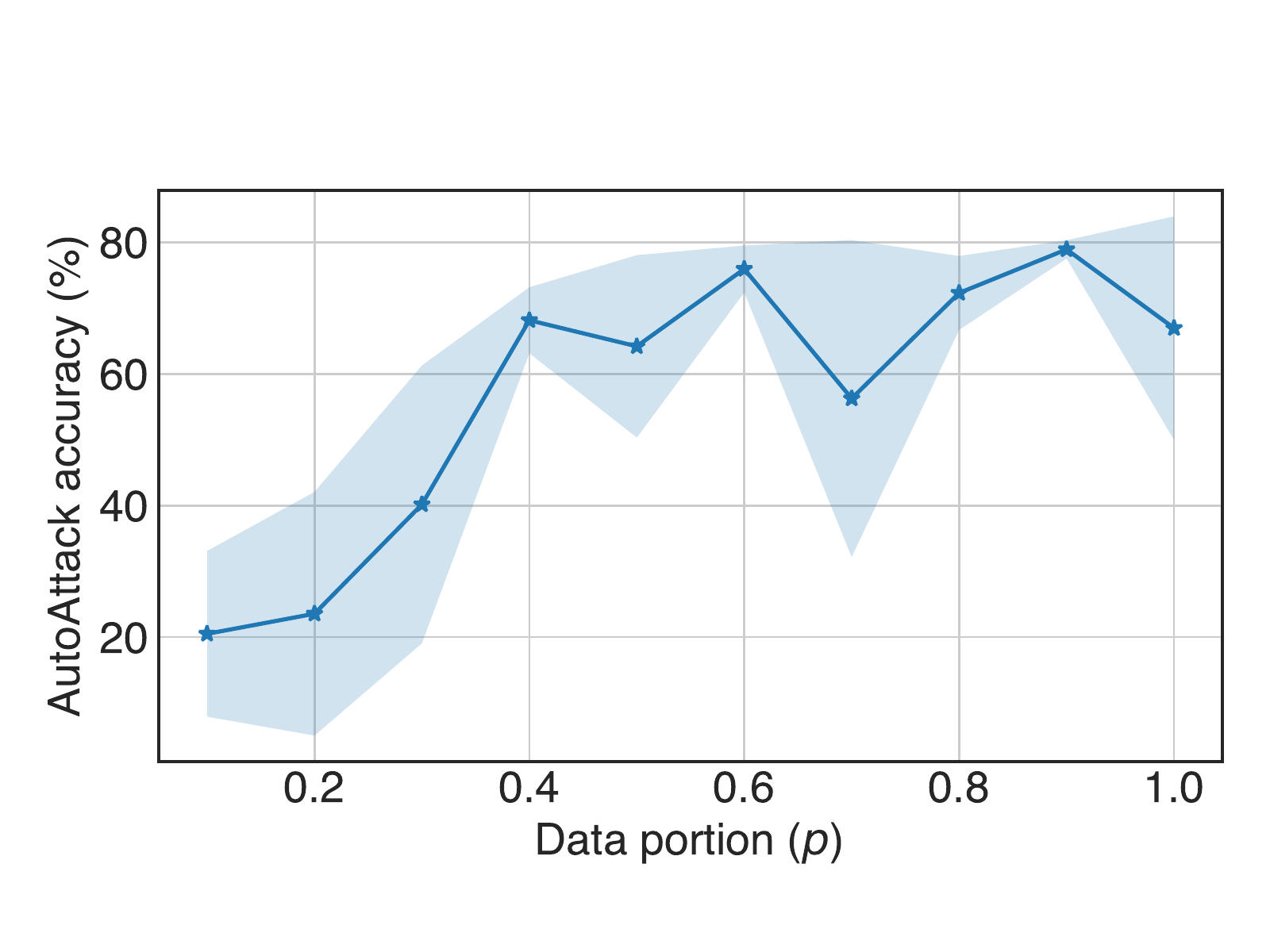}
		\caption{MNIST}
	\end{subfigure}%
	\qquad
	\begin{subfigure}{.45\textwidth}
		\centering
		\includegraphics[scale=0.4]{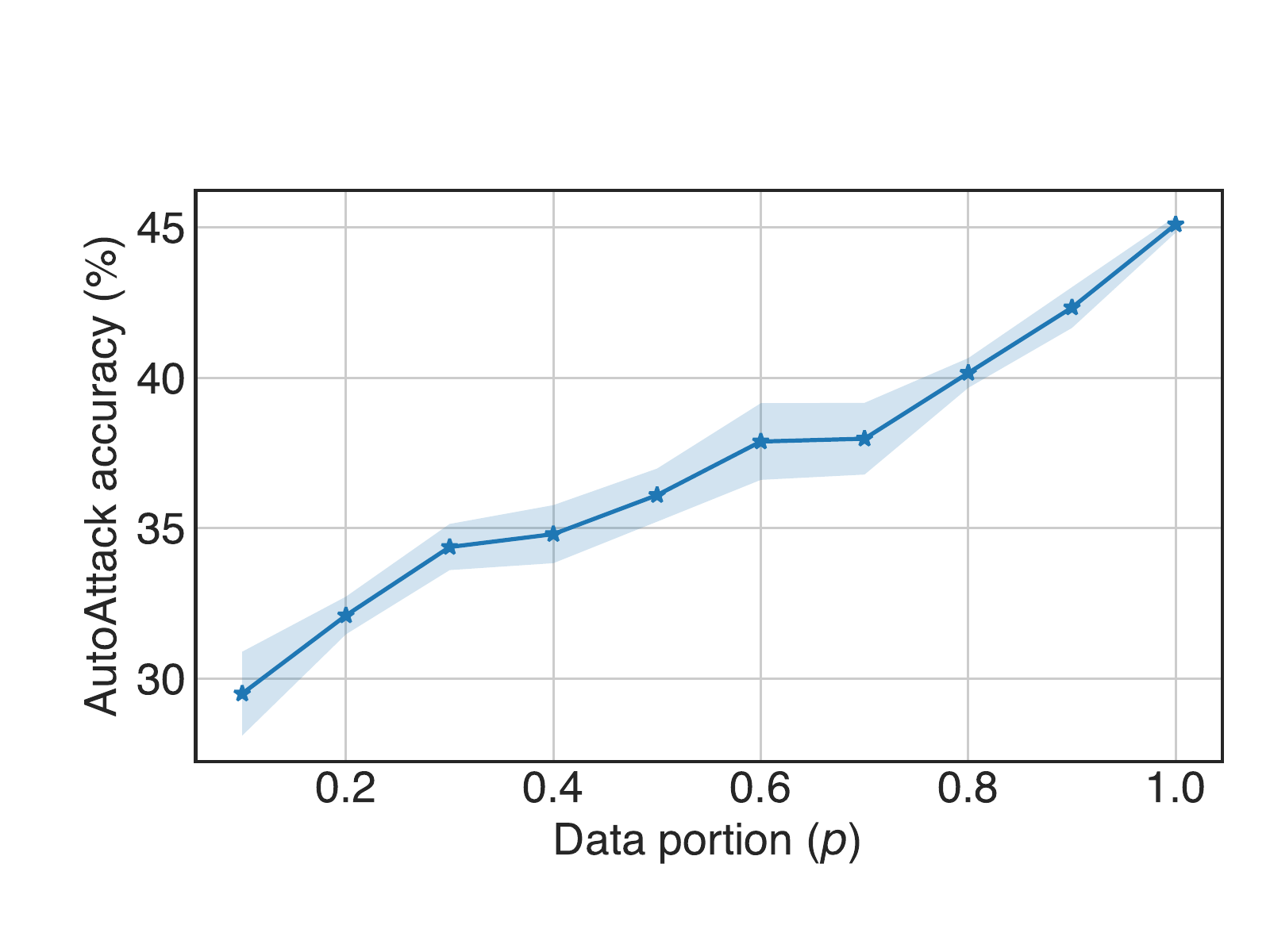}
		\caption{CIFAR10}
	\end{subfigure}
	\caption{Adversarial accuracy against the proportion of the original data ($p$) left after the dropout epoch ($e = 1$), for both MNIST and CIFAR10. Error bars denote one standard deviation, and ($n = 5$) models are plotted per datapoint.}
	\label{fig:eval_random_dropout_portion}
\end{figure}

The best proportion of data to prune appears to be case-specific. The aim is to prune as much data as possible without sacrificing the adversarial performance by a significant amount, leading to a tension between model performance and computational efficiency, a trade-off which is explored in \Cref{fig:eval_random_dropout_portion}. It is clear that the pruning proportion of the data has a significant impact on model robustness. This trend is more pronounced in the case of CIFAR10---our hypothesis for the cause of this is discussed in \Cref{discussion}. 

\textbf{\textit{Takeaway}}: \textit{Pruning proportion affects model robustness with a efficiency-utility trade-off.}

\subsection{Loss-based data pruning}
In addition to the simple strategy of randomly dropping data, we can also use loss as a guide to prune data. As explained in \Cref{fig:motivation} and \Cref{sec:method:data_prune}, high-loss data points might interfere with the model's ability to learn, and low-loss data points are points that the model are confident on. In loss-based sub-sampling, we evenly drop low-loss and high-loss data points.

\Cref{tab:compare} demonstrates how loss-based sub-sampling compares to both random data sub-sampling and when the full dataset is present. Each of the data entry is an averaged reading from 30 independent runs. We show the two different sub-sampling strategies (random and loss-based respectively) with two different pruning rates ($p=0.6, p=0.2$). 

One interesting phenomenon is that, \textit{within the MNIST dataset, a pruning rate of $0.6$, corresponds to a robustness increase of $0.06\%$.}. We also demonstrate more results on different data proportions in \Cref{fig:eval_both}. Both results demonstrate an interesting phenomenon, where the loss-based sub-sampling is better on MNIST but is worse than random pruning on CIFAR10. We will discuss further the cause of this in \Cref{discussion}.

\begin{table}[h]
	\centering 
	\caption{Performance of RS+FGSM training on MNIST and CIFAR10.
    The values in each column correspond to a single standard deviation, with $n=30$ models trained on each dataset. Time is measured in minutes.}
	\label{tab:compare}
	\adjustbox{width=\linewidth}{
	\begin{tabular}{l cc cc c}
	\toprule
	\textbf{Metric} 
	& \multicolumn{2}{c}{\textbf{Loss-based}}
	& \multicolumn{2}{c}{\textbf{Random}}
	& \textbf{Full data} \\
	& $p=0.6$
	& $p=0.2$
	& $p=0.6$
	& $p=0.2$
	\\
	\midrule
    & \multicolumn{5}{c}{\textit{MNIST}}
	\\
	\midrule
	\textbf{Accuracy}
	& $94.60 \pm 5.95$
	& $94.10 \pm 5.98$
	& $\mathbf{98.01 \pm 0.12}$
	& $97.78 \pm 1.00$
    & $97.67 \pm 0.01$
	\\
	\textbf{AA}
	& $\mathbf{70.11 \pm 7.81}$
	& $67.43 \pm 9.80$
	& $65.50 \pm 0.013$
	& $60.77 \pm 11$
	& $70.05 \pm 7.29$
	\\ 
	\textbf{Time (m)}
	& $0.51 \pm 0.01$
	& $0.74 \pm 0.02$
	& $0.46 \pm 0.01$
	& $0.69 \pm 0.02$
	& $0.78 \pm 0.02$
	\\
	\midrule
    & \multicolumn{5}{c}{\textit{CIFAR10}}
	\\
	\midrule
	\textbf{Accuracy}
	& $50.64 \pm 2.09$
	& $70.71 \pm 1.42$
	& $64.04 \pm 1.48$
	& $\mathbf{73.26 \pm 0.85}$
	& $79.68 \pm 0.27$
	\\
	\textbf{AA}
	& $25.32 \pm 1.43$
	& $37.84 \pm 1.11$
	& $33.54 \pm 1.30$
	& $\mathbf{39.63 \pm 0.62}$
	& $44.59 \pm 0.27$
	\\
	\textbf{Time (m)}
	& $16.57 \pm 0.67$
	& $26.43 \pm 1.58$
	& $\mathbf{15.96 \pm 0.39}$
	& $25.58 \pm 0.84$
	& $30.16 \pm 0.86$
	\\
	\bottomrule
	\end{tabular}
	}
\end{table}

\textbf{\textit{Takeaway}}: \textit{Loss-based data sub-sampling demonstrates effectiveness on MNIST by showing better AutoAttack accuracy than full data participation}. 

\subsection{A deep dive to loss-based data pruning}

	\begin{figure}[!h]
	\centering
	\begin{subfigure}{.45\textwidth}
		\centering
		\includegraphics[width=\linewidth]{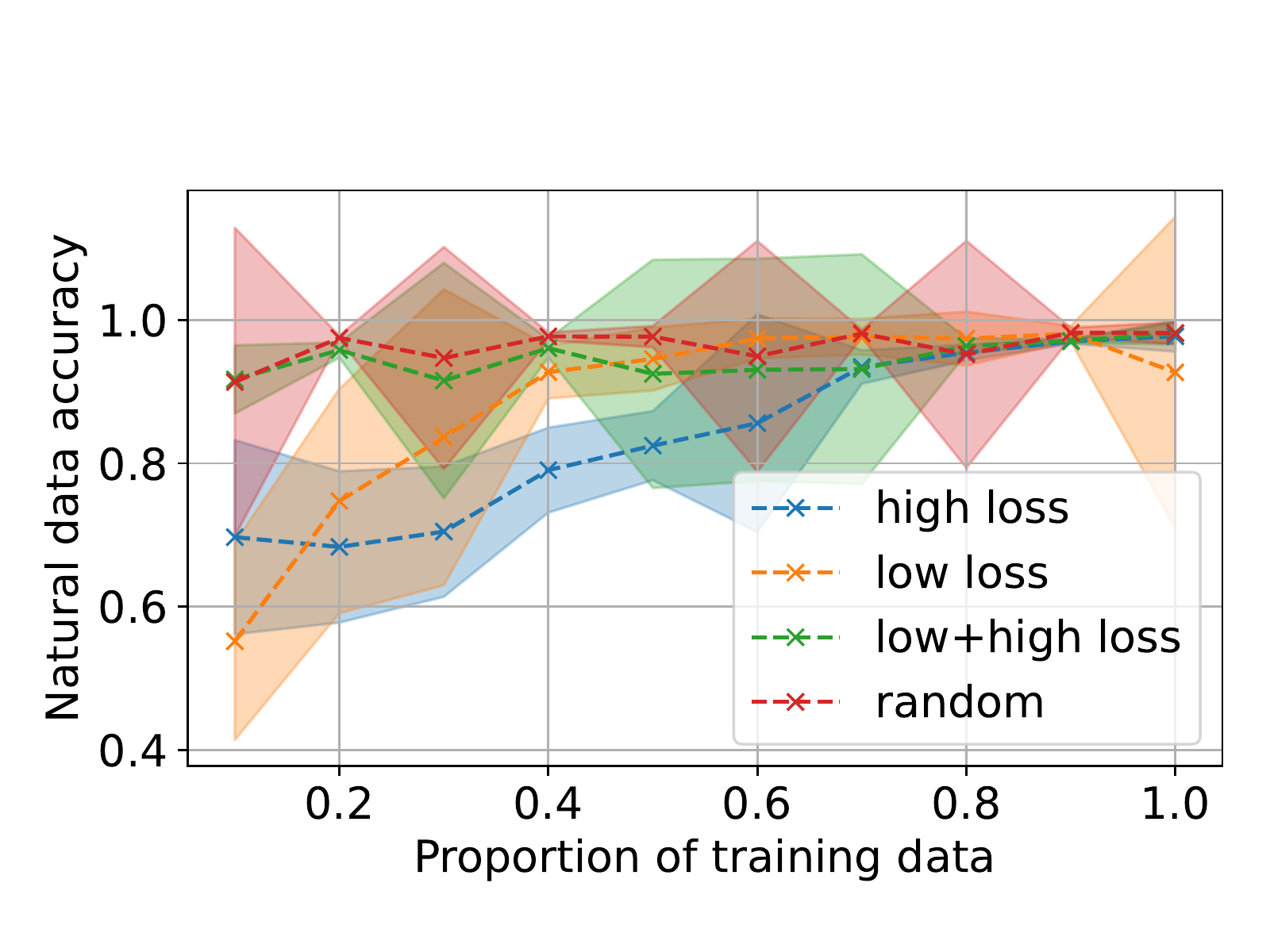}
		\caption{Accuracy}
	\end{subfigure}%
	\qquad
	\begin{subfigure}{.45\textwidth}
		\centering
		\includegraphics[width=\linewidth]{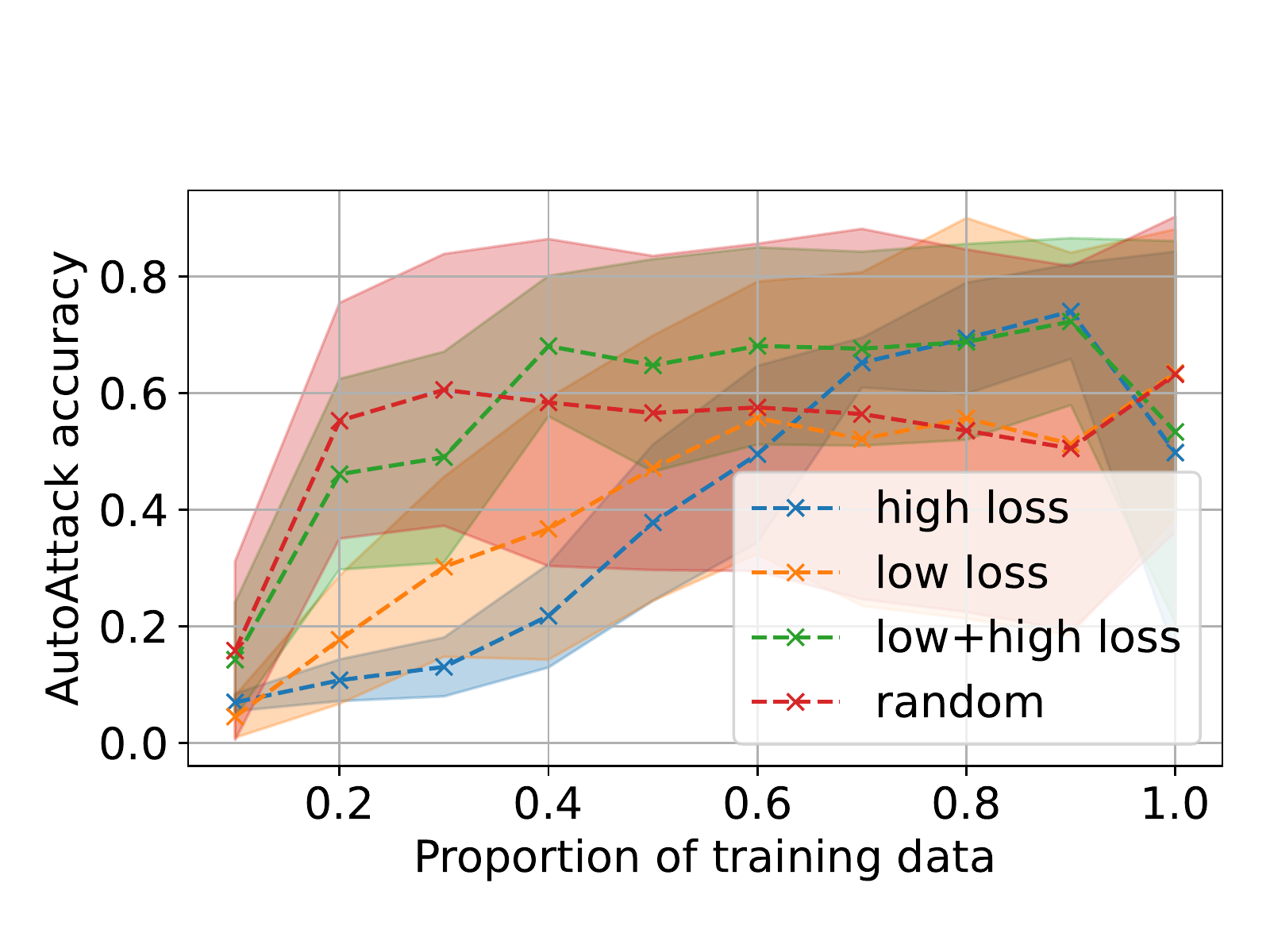}
		\caption{AutoAttack performance}
	\end{subfigure}
	\caption{Different data pruning strategies for MNIST. The horizontal axis is the proportion of training data used for adversarial training, after the pruning has been performed. We show both the test accuracy on natural test data and AutoAttack performance. Low and high (low+high) loss based sub-sampling not only maintains natural data accuracy but also gains better adversarial robustness at relatively high pruning rates, when compared to training on the full dataset.}
	\label{fig:mnist_timeperf}
	\end{figure}

\begin{figure}[!h]
	\centering
	\begin{subfigure}{.45\textwidth}
		\centering
		\includegraphics[width=\linewidth]{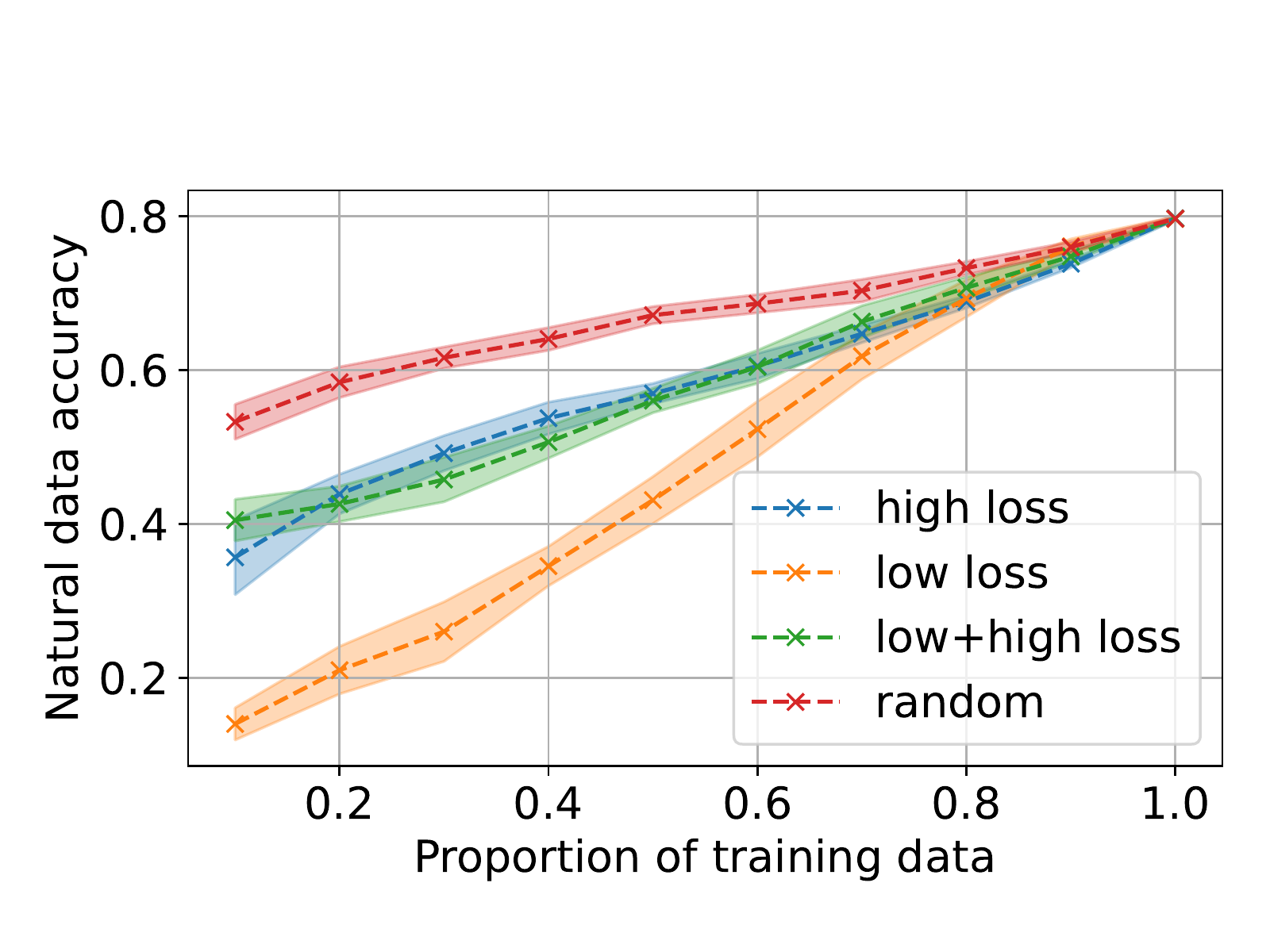}
		\caption{Accuracy}
	\end{subfigure}%
	\qquad
	\begin{subfigure}{.45\textwidth}
		\centering
		\includegraphics[width=\linewidth]{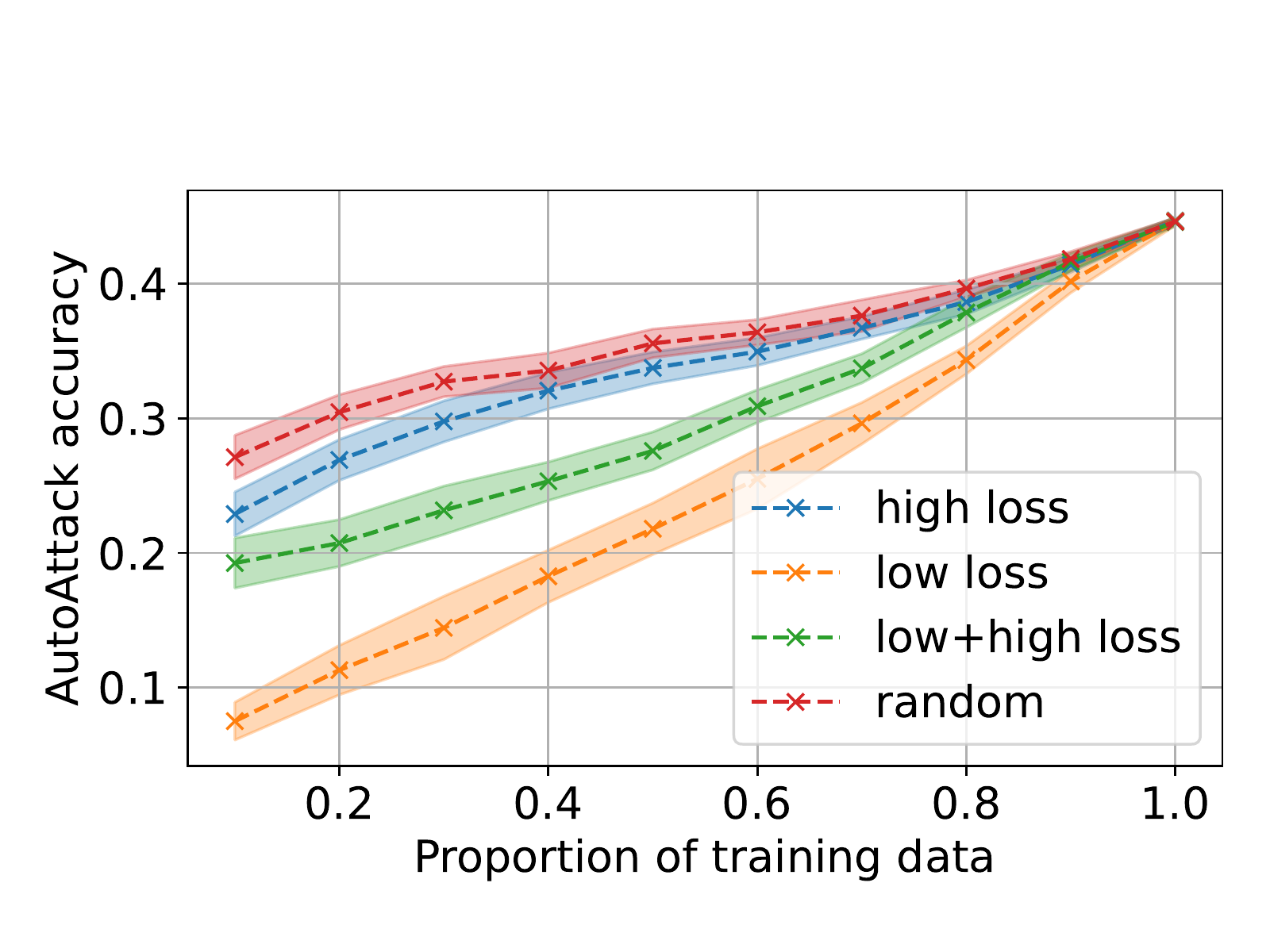}
		\caption{AutoAttack performance}
	\end{subfigure}
	\caption{Different data pruning strategies for CIFAR10. The horizontal axis is the proportion of training data used for training. We show both natural data accuracy and AutoAttack performance. Low-loss based sub-sampling is disruptive for both natural data accuracy and adversarial robustness.}
	\label{fig:cifar_timeperf}
\end{figure}

In the previous subsection we demonstrated that adversarial robustness of a given model is a function of data used, and one can speed up training by using less data with minimal detrimental impact on performance. In this section we turn to a more fine-grained analysis on what enables it and instead focus on a large scale CIFAR10 dataset. 

\Cref{fig:cifar_timeperf} shows performance of data pruned CIFAR10 model using four different pruning strategies. This figure reports both the accuracy performance on the benign test data and robustness under the strongest Auto-PGD (PGD with automatic hyperparemeter tuning) based attacker~\cite{autoattack}. 

\textbf{Random sub-sampling:} We use random sampling as a baseline to assess what causes performance degradation for each of the policies. 

\textbf{Low-loss data sub-sampling:} When dropping low-loss data points, we practically drop points in which the models are confident, intuitively corresponding to regions of the highest density. Here, we observe that when omitted early in training, low-loss data points have a detrimental impact on final model accuracy. We observe an extremely strong degradation in performance for both benign and adversarial data.

\textbf{High-loss data sub-sampling:} On the opposite side of the loss spectrum we have data with the highest loss magnitudes. These usually correspond to data that the model is least confident about, data points from the tail of the underlying distribution. When omitted on the third epoch of training, we observe that such data does impact both benign benign and adversarial accuracy, but it predominantly affects benign data performance. Indeed, we observe only a slight performance degradation when comparing to random sub-sampling. 

\textbf{Low and high loss data sub-sampling:} When sub-sampled proportionally equally, we observe that impact lands itself between low and high data sub-sampling regimes, matching benign performance of high tail sub-sampling and significantly outperforming low data sub-sampling for adversarial data. 

\textit{\textbf{Takeaway:} Experiments in this subsection indicate that}:
\begin{itemize}
    \item \textit{Adversarial performance is only marginally affected by dropping out data points in the tails of the distribution, with larger effects on standard accuracy.}
    \item \textit{It is important to preserve low loss data points, as its presence is essential for both natural data accuracy and adversarial robustness.}
\end{itemize}

\begin{figure}[!h]
	\centering
	\begin{subfigure}{.45\textwidth}
		\centering
		\includegraphics[scale=0.4]{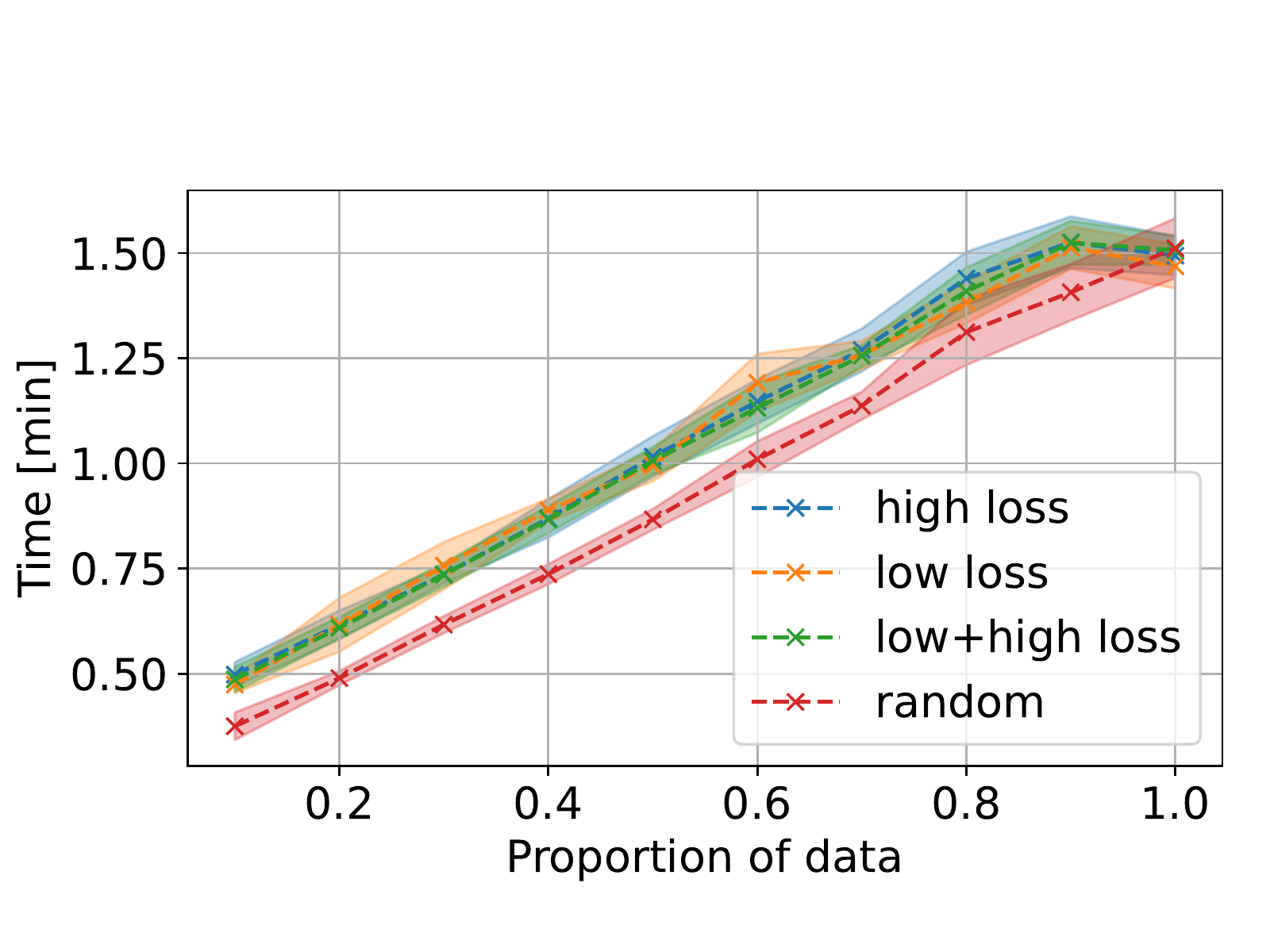}
		\caption{MNIST}
		\label{fig:mnist_training_time}
	\end{subfigure}%
	\qquad
	\begin{subfigure}{.45\textwidth}
		\centering
		\includegraphics[scale=0.4]{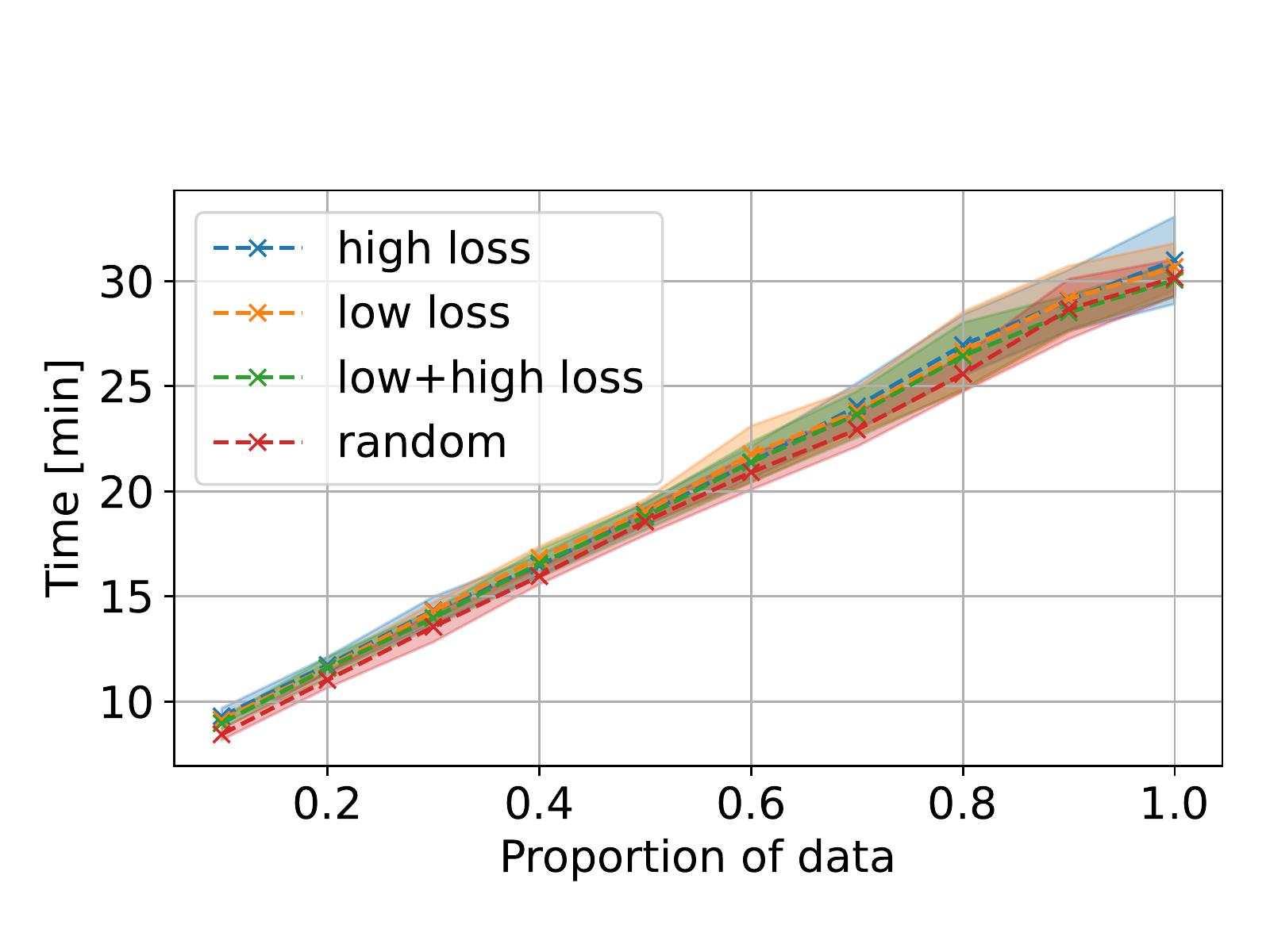}
		\caption{CIFAR10}
		\label{fig:cifar_training_time}
	\end{subfigure}
	\caption{Training time for different proportions of training data. Data pruning provides linear reduction in training time. Error bars are standard deviations computed over 30 models.}
	\label{fig:training_times}
	\vspace{-10pt}
\end{figure}

\subsection{Time savings}

In this section we turn to an empirical evaluation of the time savings from training on less data. In \Cref{sec:computational_cost} we outlined that training time complexity scales as $\mathcal{O}(NME)$, more precisely it scales linearly with the size of the dataset $M$. \Cref{fig:cifar_training_time} shows measured time requirements as a function of $M$. Here, we observe that in practice it scales linearly with data proportions. For CIFAR10 full data training takes about 30 minutes, data pruning reduces this time to approximately 18 minutes when using 50\% of the data. We further observe that most time variance occurs for the high data regimes and reduces for low data regimes. We hypothesise that this variance comes from increased memory movement. 
\section{Discussion}
As described previously, we find that the pruning of high-loss data points leads to improved performance, when compared to the removal of low-loss data. Interestingly, we also find that high-loss pruning disproportionately affects the standard accuracy of our models, while having only a marginal effect on their robust accuracy. In this section, we offer explanations for these two phenomena.

\subsection{Loss as an outlier detection mechanism}

\begin{figure}[tb]
\label{fig:pruning_visualistion}
	\centering
	\begin{subfigure}[t]{0.48\textwidth}
		\includegraphics[width=\linewidth]{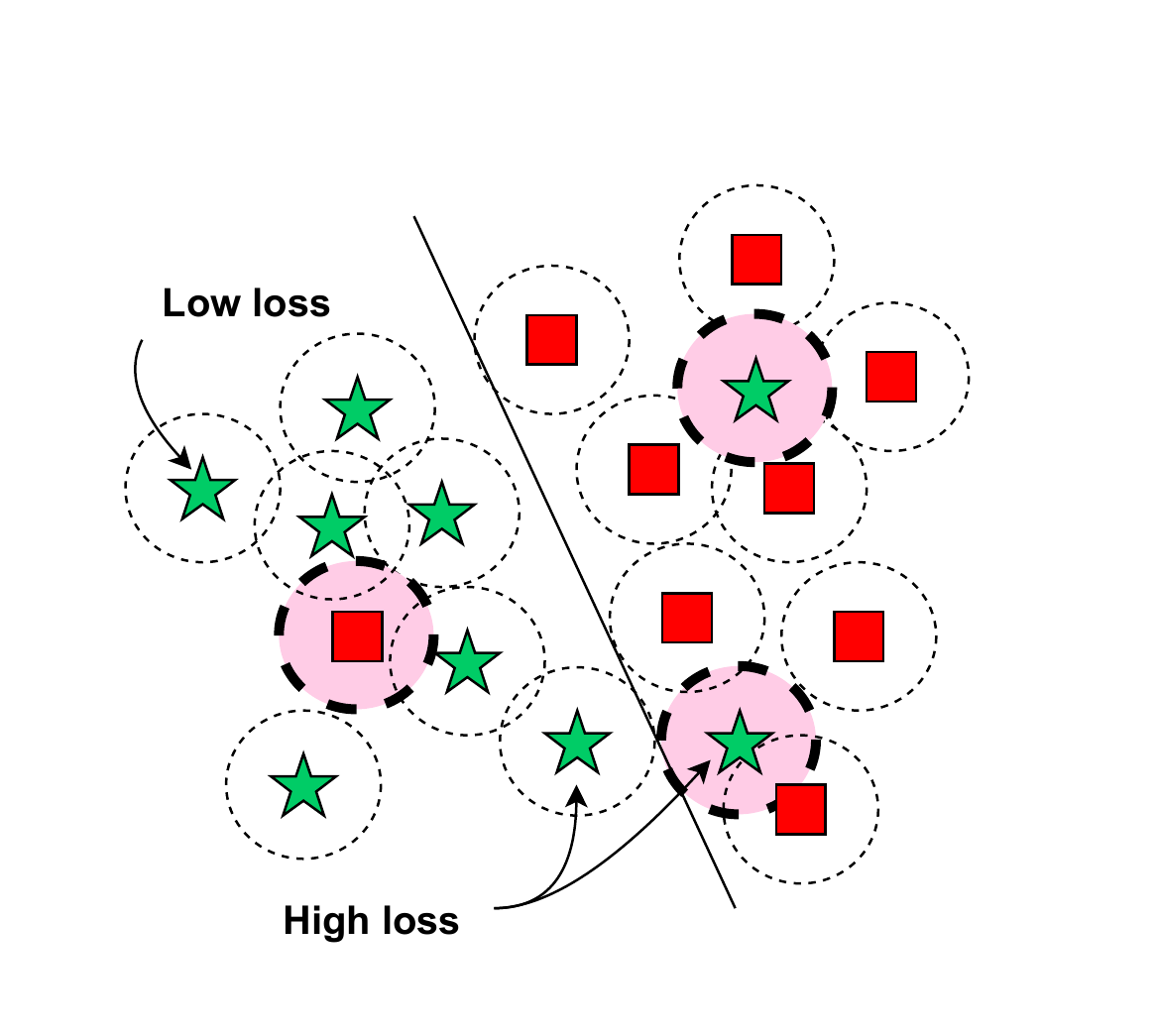}
	\caption{Noisy data crossing the decision boundary can have a negative impact on adversarial training, which is exasperated by low-loss data dropout.}
		\label{fig:dirty_example}
	\end{subfigure}~
	\begin{subfigure}[t]{0.48\textwidth}
		\includegraphics[width=\linewidth]{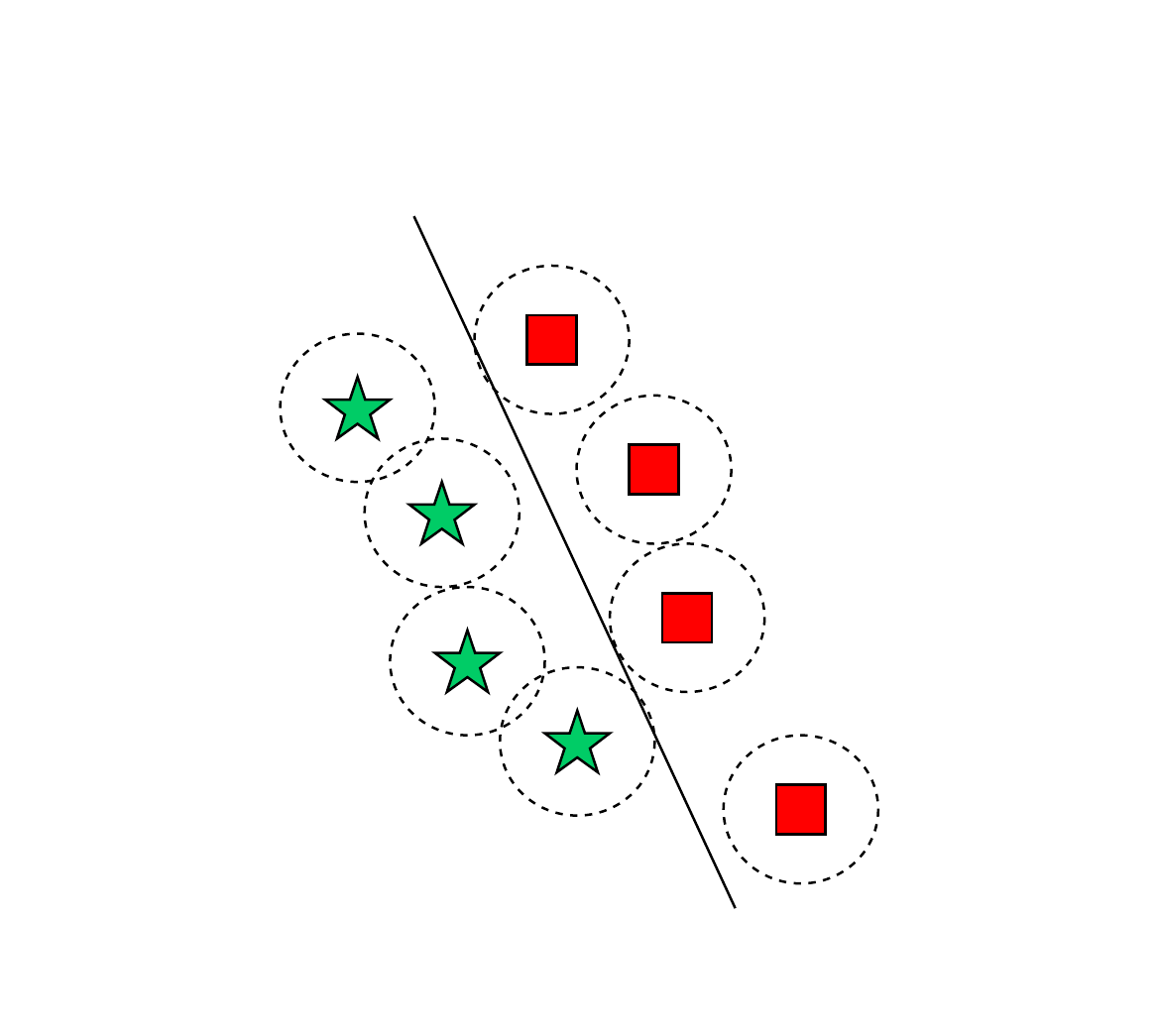}
		\caption{The less outliers we have, the clearer the decision boudnaries are, and the better training continues.}
		\label{fig:clean_example}
	\end{subfigure}
	\caption{A graphical illustration showing the issues caused by outliers during adversarial training.}
	\label{fig:motivation}
\end{figure}

In line with the existing literature~\cite{loss_based_denoising_1,loss_based_denoising_2}, we posit that the removal of high-loss datapoints corresponds to the removal of outliers within our datasets. This is particularly relevant for the datasets used in modern machine learning tasks, that contain high proportions of mislabelled datapoints~\cite{mislabelled_datasets,label_cleaning}. As illustrated in \Cref{fig:dirty_example}, we would therefore expect that the removal of low-loss data would exasperate this problem, leading to increased noise in the dataset, and hence an observed reduction in model performance. For an empirical investigation of the effects of high-loss dropout on the data distribution, see \Cref{appendix:dataset}.
\label{discussion}
\subsection{Toy model for robust features}
It was previously shown that the robustness of a model is highly linked to its ability to learn \emph{robust features}, and that this is dependant on which features are most usefully predictive within the dataset~\cite{bugsfeatures}. We present a natural toy model (inspired by \cite{robustness_odds_accuracy}), in which high loss datapoints correspond to those which are well predicted by fragile features, explaining why high-loss dropout disproportionately affects the standard accuracy of our models.

\textbf{Binary classification} The data model consists of input-label pairs $(\mathbf{x},y)$, sampled from a a distribution $\mathcal{D}$, containing both points which are well predicted by robust features and points which are well described by fragile features:
\begin{equation}
    (\mathbf{x},y) \gets \begin{cases}
        (\mathbf{x},y) \sim \mathcal{D}_\text{robust}&\text{w.p. }p_\text{R}\\
        (\mathbf{x},y) \sim \mathcal{D}_\text{fragile} &\text{w.p. }1-p_\text{R}
      \end{cases}
\end{equation}
      
Where our inputs are feature vectors, $\mathbf{x} = [x_1,\ldots ,x_{d+1}]$ and we
have that, for $L \in \{\text{robust},\text{fragile}\}$, if $(\mathbf{x},y) \sim \mathcal{D}_\text{L}$:
\begin{equation}
\begin{split}\label{eq:model}
y \stackrel{u.a.r}{\sim} \{-1, +1\},\qquad
&x_1 = \begin{cases}
        +y, &\text{w.p. }p_\text{L}\\
        -y, &\text{w.p. }1-p_\text{L}
      \end{cases},
      x_2,\ldots,x_{d+1} \stackrel{i.i.d}{\sim} \mathcal{N}(\eta y, 1).
\end{split}
\end{equation}

Where $x_1$ is a robust feature, and $x_2\ldots ,x_{d+1}$ are `fragile' features. We choose $\eta$ large enough that a simple linear classifier can attain high standard accuracy (>99\%), for which it suffices to set $\eta = 3/\sqrt{d}$. Furthermore, robust features are better correlated to the label for datapoints which are well described by robust features, i.e. $0.5 < p_\text{fragile} < p_\text{robust}$.

We can then define a simple linear classifier $f$ on this data, described by: 
\begin{equation}\label{eq:linear_classifier}
    f_{\text{pred}}(\mathbf{x}) := \operatorname{sign}(\mathbf{w}^\top \mathbf{x}),\quad
        \text{ where } \mathbf{w} :=
            [w_1,w_2,\ldots,w_{d+1}].
\end{equation}

\textbf{Robust classifiers} One finds that (see \cite{robustness_odds_accuracy}), in the absence of an adversary a classifier relying purely on fragile features (i.e. where $\mathbf{w} = [0,\dfrac{1}{d},\ldots,\dfrac{1}{d}])$ reaches high accuracy, however in the face of an adversary we find that the accuracy of this classifier drops rapidly. Indeed, for an adversary with a budget of $\epsilon \geq 2\eta$,

\begin{equation}
\begin{split}
\min_{\|\delta\|_\infty\leq \epsilon}\mathbb{P}
[f_\text{pred}(\mathbf{x}+\delta)=y] 
&=  \min_{\|\delta\|_\infty\leq \epsilon}\mathbb{P}\left[\frac{y}{d}\sum_{i=1}^{d}\mathcal{N}(\eta y, 1) - \delta >0\right] \\
    &\leq \mathbb{P}\left[\mathcal{N}\left(\eta, \frac{1}{d}\right)-\epsilon >0\right] \\
&\leq \mathbb{P}\left[\mathcal{N} \left (-\eta, \frac{1}{d}\right)>0 \right].
\end{split}
\end{equation}

This corresponds to a classification accuracy of less than 1\%. On the other hand, a robust classifier $\mathbf{w} = [1,0,\ldots,0] $, reaches an accuracy of $p_\text{R} p_\text{fragile} + (1-p_\text{R}) p_\text{robust}$, bounded from below by 50\%.

\textbf{Loss-based dropout} The above discussion suffices to motivate the idea that we are likely to receive a classifier which focuses on robust features, after adversarial training. However, in the high-loss pruning case we then run such a classifier on standard data, and remove data points which have a high loss---what does our model have to say about this?

Assuming we have received a robust classifier ($ \mathbf{w} = [1,0,\ldots,0] $) from adversarial training on a dataset $D = {(\mathbf{x}^1,y^1),\ldots,(\mathbf{x}^M,y^M)}$, and then the high loss data points are dropped out, using the typical mean squared error loss function: $\mathcal{L}(f,\mathbf{x},y) = |f(\mathbf{x}) - y|^2  = |\mathbf{w}^T \mathbf{x} - y|^2 = |x_1 - y|^2$.

We can then see that the data points $(\mathbf{x},y)$ which are dropped out are the ones in which the value of $x_1$ is different to the value of $y$. This occurs with probability $1-p_\text{robust}$ if $\mathbf{x} \in \mathcal{D}_\text{robust}$, and probability $1-p_\text{fragile}$ if $\mathbf{x} \in \mathcal{D}_\text{fragile}$. As we have that $1-p_\text{robust} < 1-p_\text{fragile}$, we can see that given two data points from $\mathcal{D}_{\text{fragile}}$ and $\mathcal{D}_{\text{robust}}$, we preferentially drop those from $\mathcal{D}_{\text{fragile}}$, from which it is harder to learn robust features. As a result, we arrive at a dataset with a higher proportion of robust features compared to the original dataset. 

This gives motivation for the empirical finding that the removal of high-loss based data points disproportionately affects the standard accuracy of a classifier, as we dropout points which are well described by fragile features. Standard accuracy is affected, as fragile features are still highly predictive of the label \cite{robustness_odds_accuracy}, but robust accuracy suffers less as datapoints enforcing robust accuracy are affected less by this dropout. In paradigms such as MNIST, where small subsets of the dataset suffice to achieve high accuracy \cite{small_data_mnist}, we may expect that the `distillation' of the dataset into points which are well predicted by robust features  may actually improve the adversarial robustness of our classifier.

Conversely, removing low-loss datapoints corresponds to  removing datapoints from $\mathcal{D}_{robust}$, leading to a dataset wherein robust features are harder for models to learn. This provides a further explanation as to why low-loss dropout was found to be degrade model performance.
\section{Conclusion}

In this paper we presented \textit{data pruning} -- a method to increase the speed of adversarial training. We show several data sub-sampling strategies for reducing the number of data points required for adversarial training. We observe an intriguing phenomenon that a loss-based sub-sampling strategy can not only reduce the training time but also improve the adversarial robustness on MNIST, hopefully opening the door for future work in this area.
\section*{Acknowledgments}

We would like to acknowledge our sponsors, who support our research with financial and in-kind contributions: CIFAR through the Canada CIFAR AI Chair, DARPA through the GARD project, Intel, Meta, NFRF through an Exploration grant, and NSERC through the COHESA Strategic Alliance. Resources used in preparing this research were provided, in part, by the Province of Ontario, the Government of Canada through CIFAR, and companies sponsoring the Vector Institute. 

\bibliography{bibliography}
\bibliographystyle{abbrv}

\appendix

\section{More results on loss-based data sub-sampling}

\begin{figure}[!h]
	\centering
	\begin{subfigure}{.45\textwidth}
		\centering
		\includegraphics[scale=0.4]{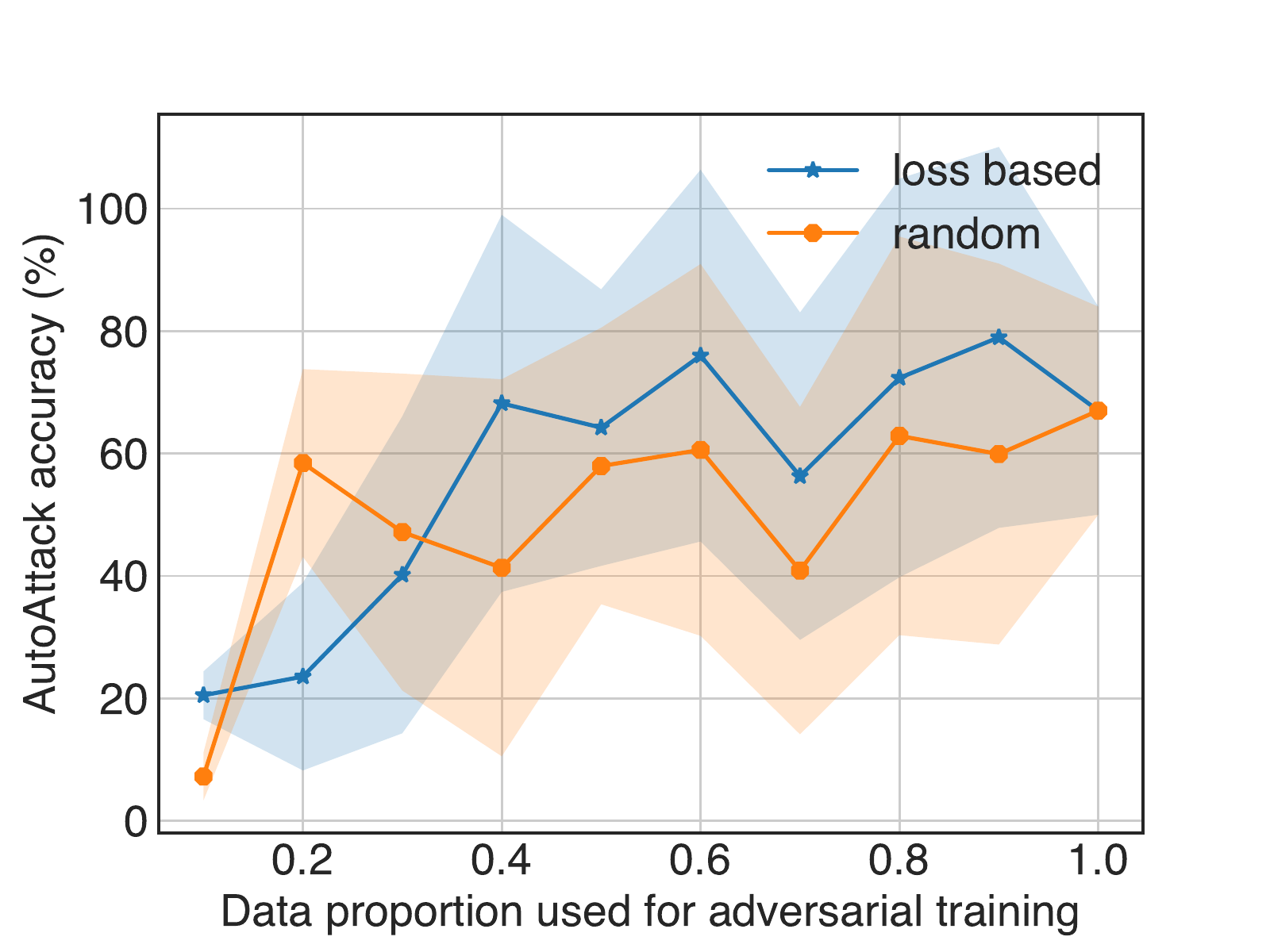}
		\caption{MNIST}
	\end{subfigure}%
	\begin{subfigure}{.45\textwidth}
		\centering
		\includegraphics[scale=0.4]{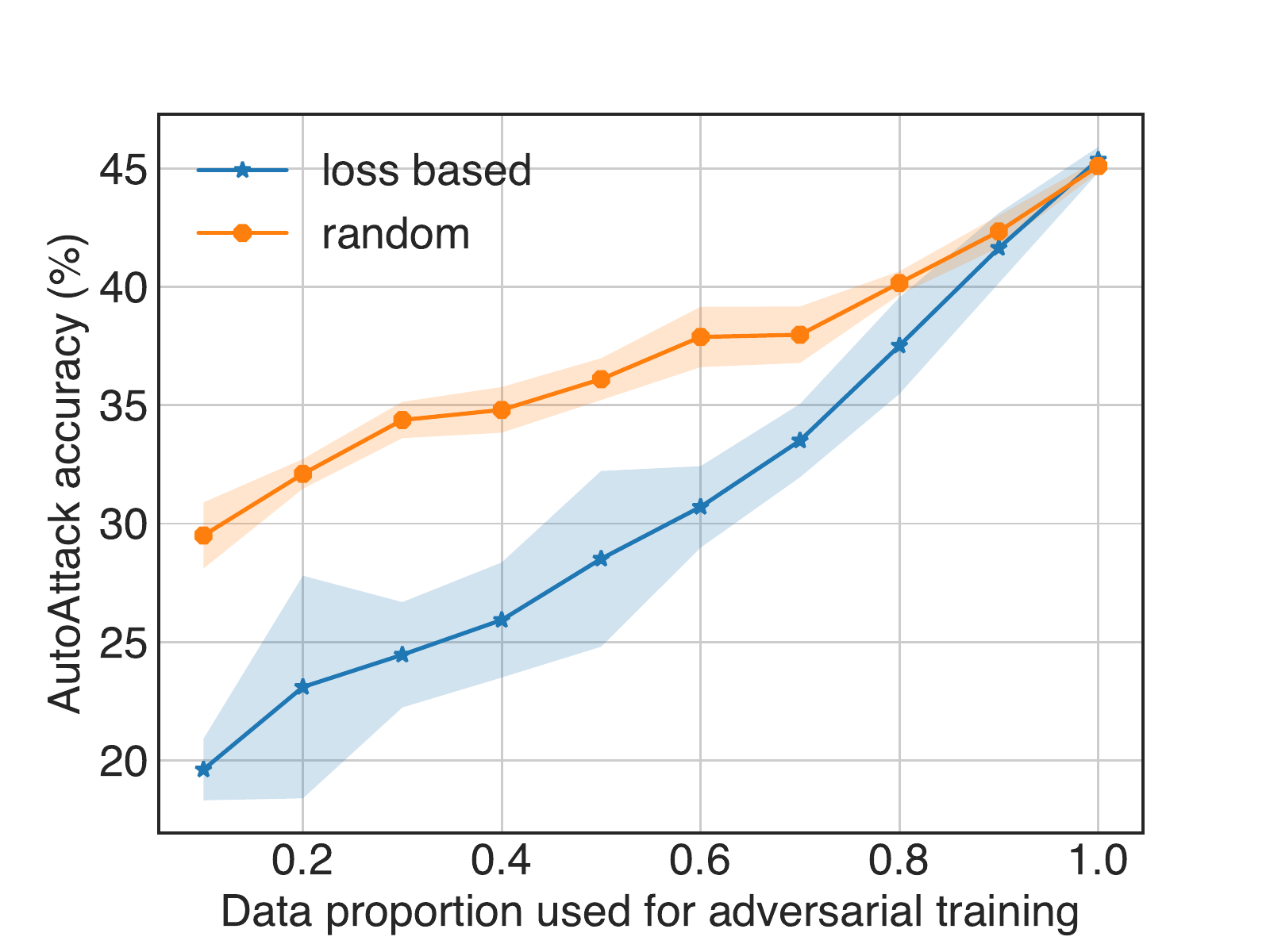}
		\caption{CIFAR10}
	\end{subfigure}
	\caption{Adversarial accuracy against the proportion of the original data ($p$) left after the dropout epoch ($e = 1$), for both MNIST and CIFAR10 and both random pruning and loss-based pruning. Error bars denote one standard deviation, and we plot ($n = 5$) models per datapoint.}
	\label{fig:eval_both}
	\end{figure}

\Cref{fig:eval_both} demonstrates an interesting phenomenon, where the loss-based sub-sampling is better on MNIST but is worse than random pruning on CIFAR10.

\subsection{Dataset statistics}
\label{appendix:dataset}
\begin{figure}[!h]
	\centering
	\begin{subfigure}{.24\textwidth}
		\centering
		\includegraphics[width=\linewidth]{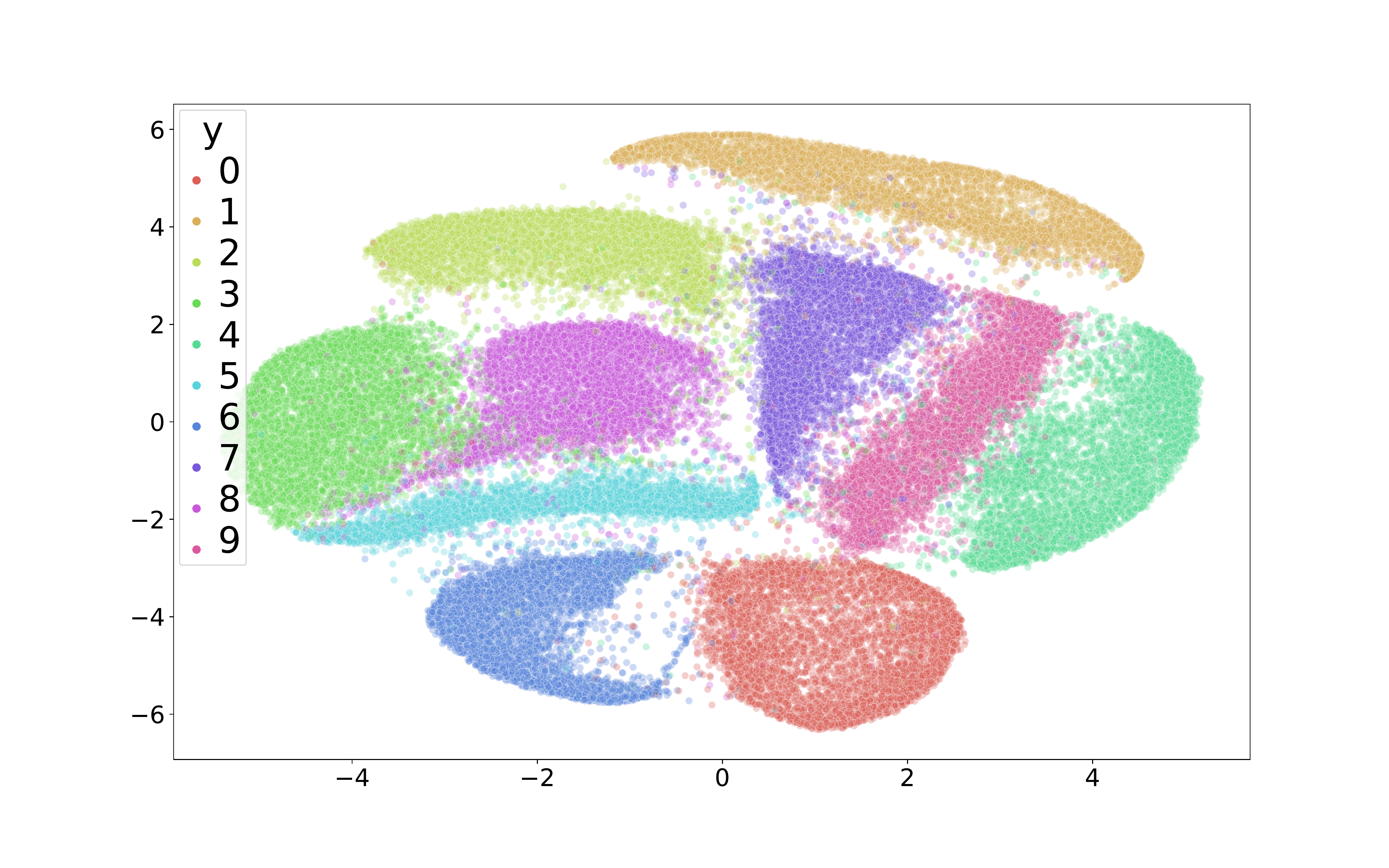}
		\caption{MNIST}
		\label{fig:dataset_illustration:mnist}
	\end{subfigure}%
	\begin{subfigure}{.24\textwidth}
		\centering
		\includegraphics[width=\linewidth]{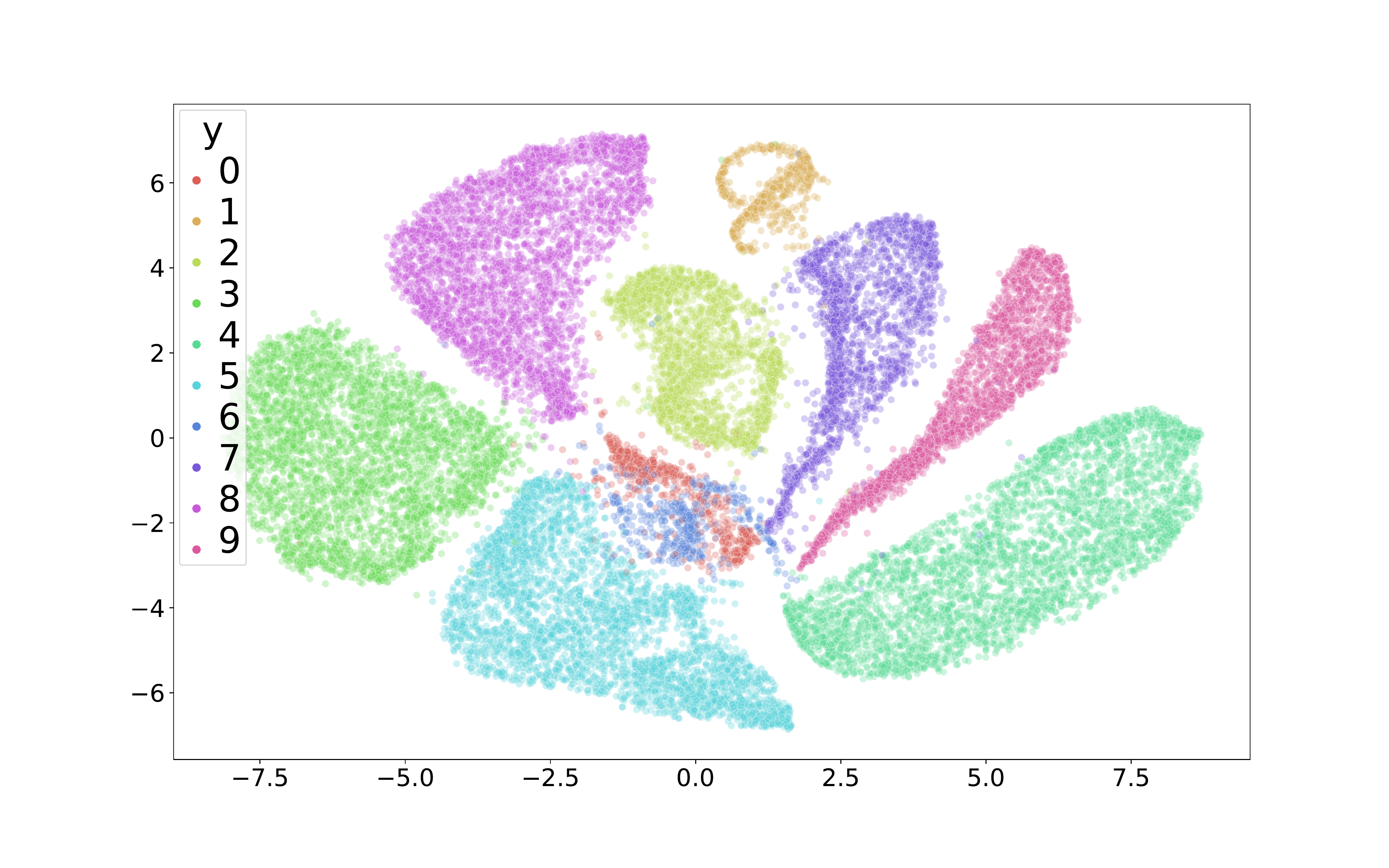}
		\caption{Pruned MNIST}
		\label{fig:dataset_illustration:pruned_mnist}
	\end{subfigure}%
	\begin{subfigure}{.24\textwidth}
		\centering
		\includegraphics[width=\linewidth]{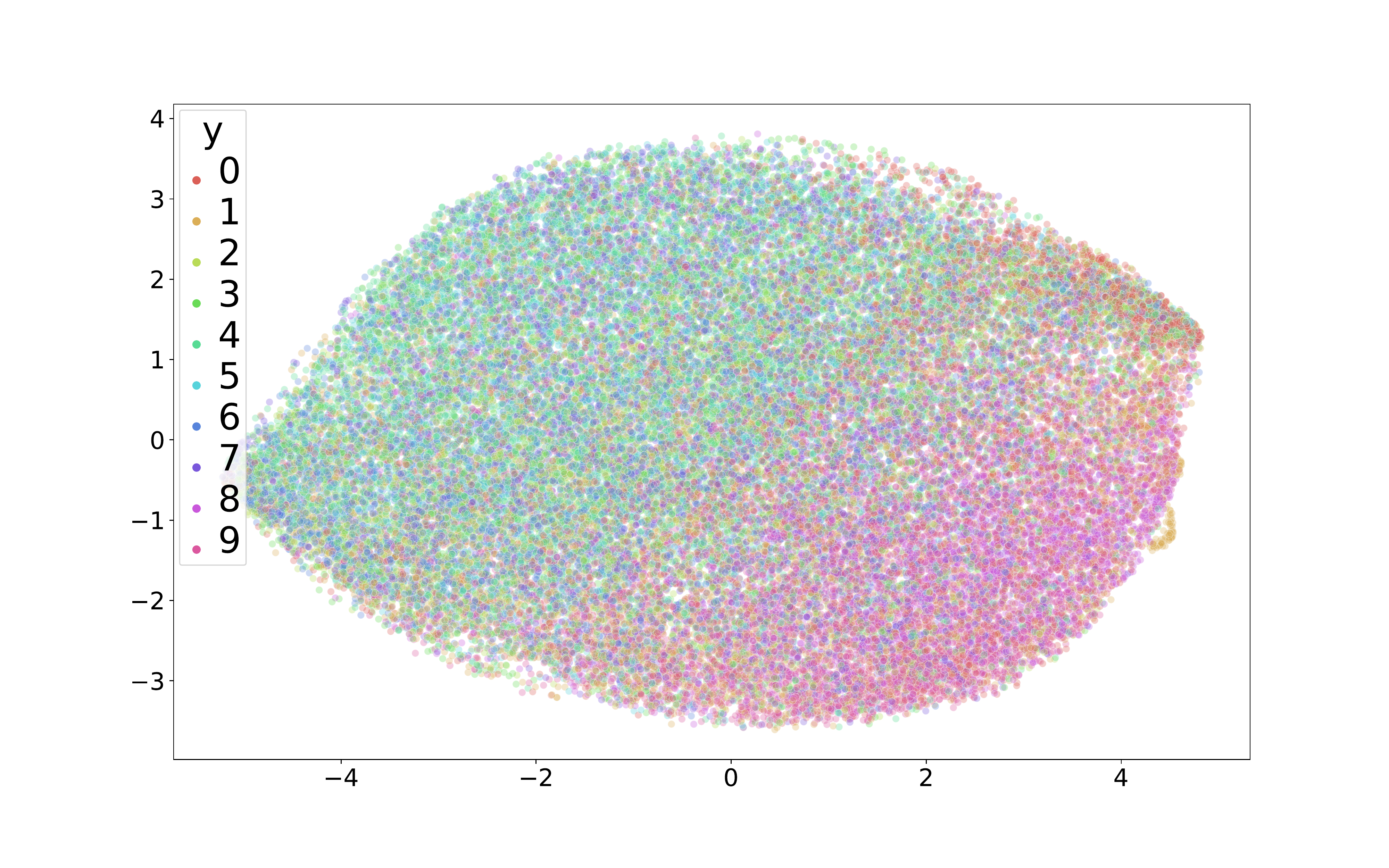}
		\caption{CIFAR10}
		\label{fig:dataset_illustration:cifar10}
	\end{subfigure}
	\begin{subfigure}{.24\textwidth}
		\centering
		\includegraphics[width=\linewidth]{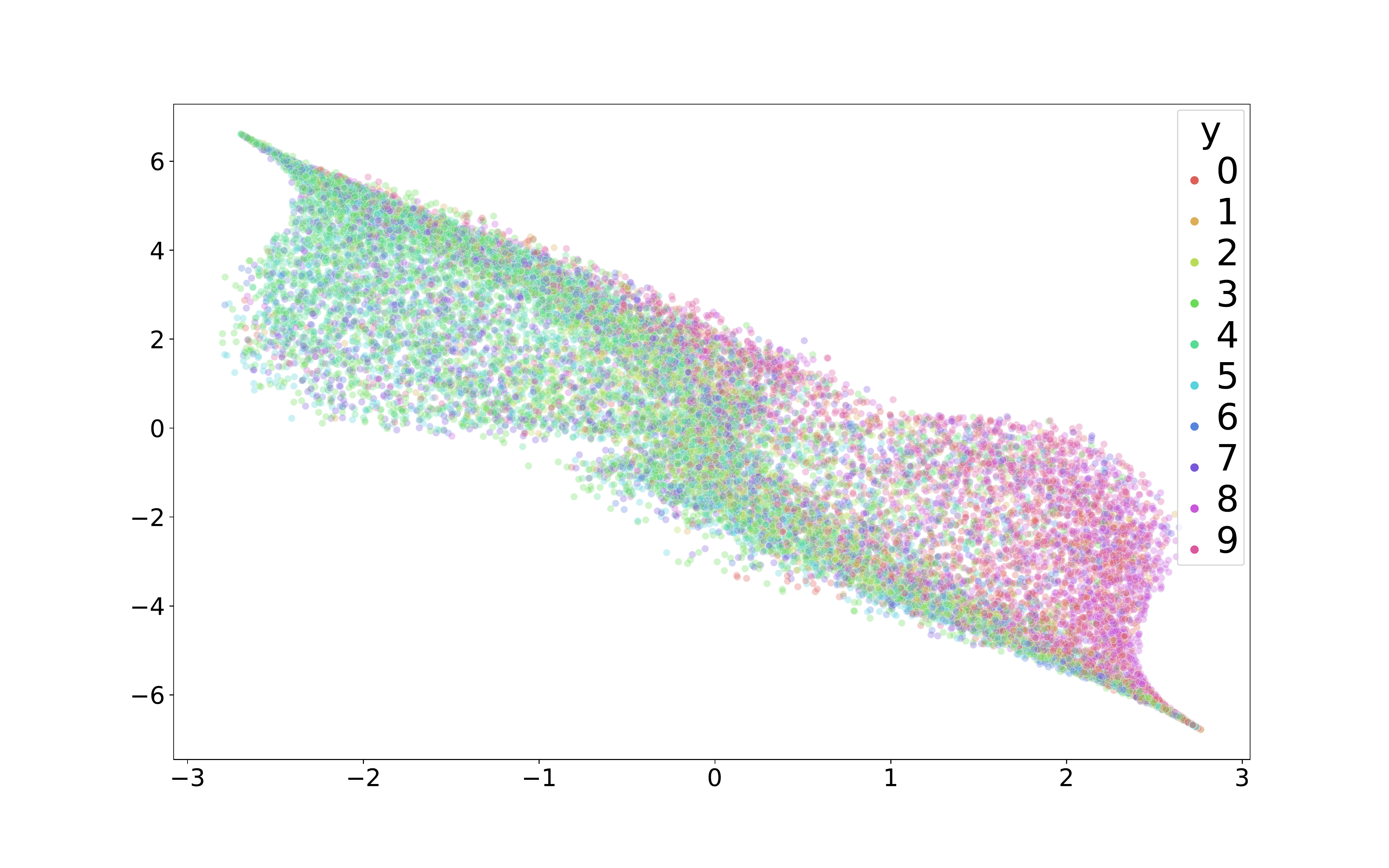}
		\caption{Pruned CIFAR10}
		\label{fig:dataset_illustration:pruned_cifar10}
	\end{subfigure}
	\caption{Visualising both MNIST, Pruned MNIST (loss-based pruning with $p=0.4$) and CIFAR10 using T-distributed Stochastic Neighbor Embedding (T-SNE) and on top PCA with 50 components.}
	\label{fig:dataset_illustration}
\end{figure}
The different impacts of loss-based data pruning on the model performance are linked to the dataset statistics. 
We first run a Principal Component Analysis (PCA) on the dataset and record its top 50 components. We then use T-distributed Stochastic Neighbour Embedding (T-SNE) \cite{van2008visualizing} with two components to visualise the datasets as shown in \Cref{fig:dataset_illustration}.

Consider a dataset $\mathfrak{D}$ with $N$ samples, $M$ classes, we use $\mathfrak{D}_m = \{(x, y) \mid y = m\ \land (x, y) \in \mathfrak{D}\}$ to represent all the data points in class $m$.
For example, $\mathfrak{D}_0$ for MNIST corresponds to the cluster for class 0 (in red in the left plot of \Cref{fig:dataset_illustration}).
We can then have the following metrics to describe the dataset:

\begin{itemize}
	\item Density ($\rho$): this metric measures the averaged density of data within each class ($\mathfrak{D}_m$), formally, we can say this is $ \rho = (\sum^{M-1}_{m=0} (\sum_{x_i \in \mathfrak{D}_m} x_i) / |\mathfrak{D}_m|)) / M$.
 \item Minimum separation ($\delta$): this metric measures the distance between a centroid and its nearest neighbour, and it is the averaged across all centroids. If a centroid $c_m = \sum_{x_i \in \mathfrak{D}_m} x_i) / |\mathfrak{D}_m|$, we then have $\delta = \sum^{M-1}_{m=0} |c_m - c'|^2$ where $c'$ is the nearest neighbour of $c_m$.
\end{itemize}

Since the effect of data noise exists for most of today's dataset, we cut off the $1\%$ data at the tails when building each $\mathfrak{D}_m$, these are data points that are far away from the centroid. This then generates us \Cref{tab:dataset_stats} that characterises our datasets.

\begin{table}[h]
	\centering 
	\caption{
		Comparing datasets using averaged class-wise density ($\rho$) 
		and minimum separation between classes ($\delta$). A larger $\rho$ and a larger $\delta$ means the dataset is easier to prune using loss-based sub-sampling.}
	\label{tab:dataset_stats}
	\begin{tabular}{cc|cc|cc}
	\toprule
	\multicolumn{2}{c}{\textbf{MNIST}}
	&\multicolumn{2}{c}{\textbf{Pruned MNIST}}
	& \multicolumn{2}{c}{\textbf{CIFAR10}} \\
	\midrule
	$\rho$
	& $\delta$
	& $\rho$
	& $\delta$
	& $\rho$
	& $\delta$
	\\
	\midrule
	$1015.55$
	& $0.7698$
	& $222.34$
	& $1.4860$
	& $96.6625$
	& $0.3571$ \\
	\bottomrule
	\end{tabular}
\end{table}

Our results in \Cref{tab:dataset_stats} confirm observations we made in previous subsections.

First, \textit{Datasets (MNIST vs. FashionMNIST vs. CIFAR10) with large $\rho$ and small $\delta$ values benefit more from pruning both low-loss and high-loss data points.}.
This is also reflected in \Cref{tab:compare} that MNIST benefits more from loss-based sub-sampling compared to CIFAR10. 
When the data points of each cluster are dense and each of these clusters are far away each other, the complexity of the dataset is lower and we can safely remove more data points, especially the low-loss and high-loss data points. This is also reflected on the phenomenon that loss-based sub-sampling is better than random sub-sampling on MNIST.

Second, \textit{Loss-based data sub-sampling will transform datasets to have lower $\rho$ and larger $\delta$}. When we look at \textbf{MNIST} and \textbf{Pruned MNIST}, we see the \textbf{Pruned MNIST} have lower $\rho$ and larger $\delta$ values.
This explain why in \Cref{tab:compare}, loss-based pruning can help MNIST to achieve better adversarial robustness compared to full data participation. If we look at 
\Cref{fig:dataset_illustration:mnist} and \Cref{fig:dataset_illustration:pruned_mnist}, they suggest that \textbf{Pruned MNIST} have clusters that are now more separated, and this would help with model robustness.





\end{document}